\documentclass{article}
\usepackage{url}			
\usepackage{booktabs}	
\usepackage{floatpag}	
\usepackage[utf8]{inputenc}	

\usepackage{amsthm}
\usepackage{amsmath}
\usepackage{amsfonts}
\usepackage{amssymb}
\usepackage{graphicx}
\usepackage{url}
\usepackage{subfigure}
\usepackage{epstopdf}
\usepackage{algorithm}
\usepackage{algorithmic}
\usepackage{subfigure}
\usepackage{fancyhdr}
\graphicspath{{../}{figures/}}

\begin{document}
\title{The Prague Relational Learning Repository}
\author{Jan Motl \\ \\ Faculty of Information Technology \\ Czech Technical University in Prague \\ Prague, Czech Republic \\jan.motl@fit.cvut.cz \and Oliver Schulte\\
\\ School of Computing Science\\ Simon Fraser University\\Vancouver-Burnaby, Canada\\oschulte@cs.sfu.ca}
\date{\today}
\maketitle

\begin{abstract}
The aim of the Prague Relational Learning Repository is to support machine learning research with multi-relational data. The repository currently contains 148 SQL databases hosted on a public MySQL server located at \url{relational.fel.cvut.cz}. The server is provided by CTU, the Czech technical university, to support the relational machine learning community. A searchable meta-database provides metadata (e.g., the number of tables in the database, the number of rows and columns in the tables, the number of self-relationships).
\end{abstract}

\section{Goals} \label{sec:goals} Many organizations maintain their data in relational databases, which support complex structured data. Extending machine learning from traditional single-table methods to multi-relational data is an important direction for practical applications. The statistical and algorithmic challenges that arise from multi-relational data have been addressed in a number of research communities, such as Statistical-Relational Learning, Multi-Relational Data Mining, and Inductive Logic Programming. Experience with the UCI Machine Learning Repository\footnote{\url{http://archive.ics.uci.edu/ml/}} has shown that a shared repository of benchmark datasets facilitates research progress \cite{Bay2000}. The UCI Machine Learning Repository contains mainly datasets stored in a single data table. Our goal is to provide a similar service for the relational learning community for relational datasets that contain multiple interrelated tables.

\section{Design} \label{sec:design}

The repository is maintained in a public MySQL server hosted by Czech  Technical University in Prague. Each dataset is stored as a MySQL database on the server.  Different formats have been introduced for storing multi-relational data. The advantages of using the SQL (SQL stands for ``Structured Query Language'') format include the following.

\begin{itemize}
\item The SQL format is a based on a standard widely used in industry. Using SQL databases in machine learning facilitates cross-community knowledge transfer and collaborations between machine learning and database researchers.
\item Because SQL is a common standard, many programming environments support accessing and processing SQL data. This includes machine learning and statistical platforms such as Clowdflows~\cite{Kranjc2012}, RapidMiner, and Weka. All general application languages provide SQL database connectivity, including R, Python, Java, and C++. 
\item The data description facilities of SQL provide a standard for defining metadata about the structure of the dataset. For example, information about the entities linked by a relationship is specified using primary and foreign keys. This metadata is recorded in the system catalog, and can be queried by machine learning applications. 
\end{itemize}

To facilitate using tools developed for other relational data formats, we have provided scripts for converting MySQL data to other common data formats used in relational learning.\footnote{\url{https://www2.cs.sfu.ca/~oschulte/jbn/DataConversion/MLN.html}} This includes the Wisconsin Logic Learning format (WILL) and the .db format used in the Alchemy system. The ClowdFlows system also provides data format conversion, for example from MySQL to the Aleph Inductive Logic Programming Format.\footnote{\url{http://www.cs.ox.ac.uk/activities/machlearn/Aleph/aleph.html}}

\section{Content}
The repository currently contains 148 databases. This includes common benchmark datasets used in relational learning, like eastbound/westbound train dataset \cite{Trains} or biodegradability dataset \cite{Biodegradability}. Some of the databases are derived from the same base data in different ways (e.g. the repositor contains different version of the IMDb dataset). We have aimed at providing a diversity of databases, for instance in terms of the number of records and in terms of the complexity of the relational schema. Hence,
 also synthetic datasets from different database vendors are included, as they are designed to show off capabilities of their database software. An example of such a synthetic dataset is Adventure Works, which is interesting not only because of its complexity, but also because of:
\begin{itemize}
  \item it uses both, simple and composite keys;
  \item it contains a diverse set of data types, including datetime, blob (images) and geometry;
  \item it contains missing values.
\end{itemize}

\section{Access and Contributions} \label{sec:access}

Read-only access can be obtained via a database connection with the following parameters.

\begin{description}
\item[Hostname] relational.fel.cvut.cz
\item[Port] 3306
\item[Username] guest
\item[Password] ctu-relational
\end{description}

To contribute a database, please contact  the repository administrators; a web contact form is available~\url{https://relational.fel.cvut.cz/contact}. One possibility is to provide us with a MySQL dump of your database. Another option is to provide us with read access to your database on your server, so we can migrate the database to the public server. 

\section{The Meta-Database} \label{sec:metadata}

Table~\ref{table:meta} shows selected metadata from the meta-database. The meaning of the columns is as follows.

\begin{description}
\item[\#Relations] The number of tables in the database.
\item[\#Instances] Count of rows in the target table.
\item[Size] Size in MB including indexes.
\item[Type] The dataset is either a measurement or synthetically generated.
\item[Domain] The original domain.
\item[Task] Classification or regression.
\end{description}

\begin{table}
\thisfloatpagestyle{empty}
    \begin{tabular}{lrrrlll}
    \toprule
    Database & \#Relations & \#Instances & Size & Type & Domain & Task \\
    \midrule
    Accidents \cite{Wordification} & 4 & 495760 & 210.0 & Real & Goverment & Class \\
    AdventureWorks & 71 & 30669 & 234.6 & Synth & Retail & Regr \\
    AustralianFootball & 4 & 3036 & 38.0 & Real & Sport & Class \\
    Biodegradability \cite{Biodegradability} & 5 & 328 & 3.2 & Real & Medicine & Regr \\
    Carcinogenesis \cite{Carcinogenesis} & 6 & 329 & 26.3 & Real & Medicine & Class \\
    CCS & 6 & 1000 & 658.4 & Real & Finance & Regr \\
    ClassicModels & 8 & 273 & 0.5 & Synth & Retail & Regr \\
    Countries & 4 & 247 & 8.6 & Real & Geography & Regr \\
    Credit & 9 & 10084 & 443.6 & Synth & Retail & Class \\
    CS & 8 & 100 & 0.3 & Synth & Finance & Class \\
    Dunur \cite{DunurElti} & 20 & 276 & 0.8 & Real & Kinship & Class \\
    Elti \cite{DunurElti} & 14 & 1081 & 0.7 & Real & Kinship & Class \\
    Employee & 7 & 2838426 & 344.6 & Synth & Retail & Regr \\
    Financial \cite{Financial} & 8 & 682 & 94.1 & Real & Finance & Class \\
    FTP & 2 & 29555 & 7.5 & Synth & Retail & Class \\
    Genes \cite{Genes} & 3 & 862 & 1.9 & Real & Medicine & Class \\
    Hepatitis & 7 & 500 & 2.2 & Real & Medicine & Class \\
    Hockey \cite{NBA} & 23 & 7759 & 15.5 & Real & Sport & Class \\
    IMDb \cite{NBA} & 7 & 794625 & 614.6 & Real & Entertainment & Class \\
    MovieLens \cite{Schulte} & 7 & 6039 & 151.9 & Real & Entertainment & Class \\
    Lahman & 25 & 23111 & 84.0 & Real & Sport & Regr \\
    LegalActs & 5 & 564268 & 238.2 & Real & Gouverment & Class \\
    Mesh \cite{Mesh} & 32 & 223 & 1.1 & Real & Industry & Regr \\
    Mondial \cite{Mondial} & 33 & 454 & 3.3 & Real & Geography & Class \\
    MooneyFamily \cite{MooneyFamily} & 72 & 92 & 3.3 & Synth & Kinship & Class \\
    Mutagenesis \cite{Mutagenesis} & 3 & 188 & 0.9 & Real & Medicine & Class \\
    Nations & 3 & 14 & 2.1 & Real & Geography & Class \\
    NBA \cite{NBA} & 4 & 30 & 0.3 & Real & Sport & Class \\
    NCAA & 10 & 268 & 40.6 & Real & Sport & Class \\
    Northwind & 29 & 830 & 1.1 & Synth & Retail & Regr \\
    Pima & 14 & 768 & 0.8 & Real & Medicine & Class \\
    PremiereLeague \cite{PremierLeague} & 4 & 363 & 11.3 & Real & Sport & Class \\
    PTE \cite{PTE} & 41 & 299 & 7.3 & Real & Medicine & Class \\
    Pubs & 11 & 18 & 0.4 & Synth & Retail & Regr \\
    Sakila & 16 & 15991 & 6.6 & Synth & Retail & Regr \\
    SalesDB & 4 & 6148886 & 539.3 & Synth & Retail & Regr \\
    SameGen & 7 & 1081 & 0.3 & Real & Kinship & Class \\
    Stats & 8 & 38357 & 621.4 & Real & Education & Regr \\
    StudentLoan \cite{StudentLoan} & 13 & 1000 & 0.9 & Real & Education & Class \\
    PTC \cite{PTC} & 4 & 343 & 7.8 & Real & Medicine & Class \\
    Thrombosis \cite{Thrombosis} & 3 & 806 & 1.9 & Real & Medicine & Class \\
    TPCC \cite{TPCC} & 9 & 28433 & 174.1 & Synth & Retail & Class \\
    TPCDS \cite{TPCDS} & 24 & 99550 & 4587.5 & Synth & Retail & Class \\
    TPCH \cite{TPCH} & 8 & 148255 & 1925.1 & Synth & Retail & Regr \\
    Trains \cite{Trains} & 2 & 20 & 0.1 & Synth & Logistic & Class \\
    University & 5 & 38 & 0.3 & Synth & Education & Class \\
    UW-CSE \cite{UW-CSE} & 4 & 278 & 0.2 & Real & Education & Class \\
    VOC & 8 & 8215 & 2.7 & Real & Logistic & Class \\
    World & 3 & 239 & 0.8 & Real & Geography & Class \\
    \bottomrule
    \end{tabular}
    \caption{List of databases in the repository}
	\label{table:meta}
\end{table}

The name of the meta-database schema is {\tt meta}. This schema contains a number of tables with information about the databases, as well as the performance of different learning algorithms on the databases. The name of the table that contains information about the databases is {\tt meta.information}. 
Some of this metadata is automatically exported in HTML format for display on the webpage \url{relational-data.org}. In the following, we list the names of the main column and their meaning. When we refer to ``all columns'' or ``all rows'', we mean all columns/rows of all tables in a database. The metadata contain the following main groups of information: basic database statistics, information about columns or fields, foreign key structure, classification information. 

\subsection{Basic Database Statistics}

Various basic properties, such as record count and missing values. 

\begin{description}
\item[row\_count] The total number of rows, or records.
\item[row\_max] The maximum number of rows, or records, in a single table.
\item[column\_count] The total number of columns, or fields.
\item[download\_url] A URL containing further information about the dataset, such as provenance.
\item[null\_count] The number of table entries with null values; typically this is the number of table entries with missing values. 
\end{description}

\subsection{Column Information}

These columns contain metainformation about the types of columns/fields/attributes in the database tables. The list is mutually exclusive and collectively exhaustive as it holds: $column\_count=geo\_count+date\_count+lob\_count+string\_count+numeric\_count$.

\begin{description}
\item[geo\_count] The number of columns that represent spatial attributes. (These are called ``geographic'' features in MySQL.)
\item[date\_count] The number of columns  that represent temporal  attributes (date, time, or  year).  
\item[lob\_count] The number of columns  that store large objects (e.g., images).
\item[string\_count] The number of columns  that store string values. This typically includes discrete attributes.
\item[numeric\_count] The number of numeric columns.
\end{description}

\subsection{Foreign Key Structure}
A foreign key points from one table to another. Chen {\em et al.} propose visualizing the foreign key relationships in a semantic relationship graph~\cite{han2009}: The graph contains a directed edge from table $T$ to table $T'$ if table $T$ references $T'$ in a foreign key constraint.
These columns represent information about the structure of the semantic relationship graph. 

\begin{description}
\item[primary\_key\_count] The number of primary keys.
\item[composite\_key\_count] The number of primary keys that comprise more than one column.
\item[foreign\_key\_count] The number of foreign keys.
\item[self\_referencing\_table\_count] The number of tables such that the table contains a foreign key pointer to one of its own columns. This occurs for example when a relational schema represents a class hierarchy or taxonomy.
\item[has\_loop] Whether there exists a loop of foreign key pointers over several tables. An example of a loop is when between a person table and a university table exists two foreign keys - the first foreign key signifies that a person is studying at a university, while the second foreign key signifies that the person is teaching at the university.
\end{description}

\subsection{Classification}

Many of the databases in the repository have been used to study classification in relational data. There is often a standard class label for such studies; we refer to this as the {\em target attribute}. These columns contain information relevant to the target attribute where it exists.

\begin{description}
\item[target\_column] The target attribute most often used in relational classification studies.
\item[target\_table] The table that contains the target column.
\item[target\_id] The primary key field of the target table.
\item[instance\_count] The number of rows in the target table.
\item[class\_count] The number of class labels.
\item[majority\_class\_ratio] The proportion of the majority class label on instance count.
\end{description}

\section{Conclusions}
In  this  paper,  we  presented  the Prague Relational Learning Repository (PRLR), an easily accessible collection of datasets for relational learning. The PRLR was designed with supervised learning in mind. To this end, the PRLR contains 148 ready to download datasets. One of the important features of the PRLR is that it provides meta-data about the datasets. The PRLR meta-data can be accessed at \url{https://relational.fel.cvut.cz}.

\section*{Acknowledgment} We are grateful to CTU Prague for sponsoring and maintaining the database server. Our industry partner getML ensured service continuity by hosting the database server during a transition phase (\url{www.getml.com}). 
We would like to thank all of the donors who contributed data to the repository, and the author of the web page, Václav Ostrožlík. This work was supported by the Grant Agency of the Czech Technical University in Prague, grant No. SGS15/117/OHK3/1T/18.

\bibliography{library}{}
\bibliographystyle{plain}

\end{document}